# TimeLesSeg: Unified Contrast-Agnostic Cross-Sectional and Longitudinal MS Lesion Segmentation via a Stochastic Generative Model

Vicent Caselles-Ballester[1], Eloy Martínez-Heras[2], Giuseppe Pontillo[3,4,5,6], Zoe Mendelsohn[3], Elena M. Marrón[1], Juan Luis García Fernández[1], Laia Subirats[1], Jon Stutters[3], Jeremy Chataway[3,7], Frederik Barkhof[3,4,6,8], Sara Llufriu[2], Ferran Prados[1,3,8]

*1 - NeuroADaS Lab, eHealth Center, Universitat Oberta de Catalunya, Barcelona, Spain.*

*2 - Neuroimmunology and Multiple Sclerosis Unit, Laboratory of Advanced Imaging in Neuroimmunological Diseases (ImaginEM), Hospital Clinic Barcelona, Institut d'Investigacions Biomèdiques August Pi i Sunyer (IDIBAPS) and Universitat de Barcelona. Barcelona, Spain.*

*3 - Queen Square Multiple Sclerosis Centre, Department of Neuroinflammation, UCL Queen Square Institute of Neurology, Faculty of Brain Science, University College of London, London, United Kingdom.*

*4 - MS Center Amsterdam, Amsterdam Neuroscience, Amsterdam UMC, Vrije Universiteit Amsterdam, Amsterdam, The Netherlands*

*5 - Departments of Advanced Biomedical Sciences and Electrical Engineering and Information Technology, University of Naples "Federico II," Naples, Italy*

*6 - Radiology & Nuclear Medicine, Amsterdam University Medical Centers, Vrije Universiteit, Amsterdam, The Netherlands.*

*7 - National Institute for Health and Care Research, University College London Hospitals Biomedical Research Centre, London, United Kingdom*

*8 - UCL Hawkes Institute, Department of Medical Physics and Bioengineering, University College London, London, UK.*

**ABSTRACT**

Multiple sclerosis (MS) expresses substantial clinical and radiological heterogeneity, which poses significant challenges for automatic lesion segmentation, a crucial process in both diagnosis and disease monitoring. In particular, deep learning methods are well-known for their poor generalization across scanners, contrasts, and rigidity to the temporal structure of inputs. As such, existing approaches are constrained to either cross-sectional or longitudinal segmentation, and struggle with changes in data distribution or when temporal information is unavailable.

We introduce TimeLesSeg, a unified contrast-agnostic framework designed to segment MS lesions regardless of the presence of a temporal dimension in its inputs, and with a single convolutional neural network. Our approach models pathological priors through lesion masks, which are processed together with the current MRI scan. Cross-sectional processing is enabled by exposing the model to training cases where no prior information is available, which are modeled with an empty mask, allowing it to operate seamlessly in both scenarios.

To overcome the scarcity and inconsistency of longitudinal datasets, we propose a novel generative pipeline, in which patterns of lesion evolution are simulated by stochastically deforming each individual lesion with morphological operations, producing realistic prior timepoints. In parallel, we achieve contrast agnosticism through Gaussian mixture model–based domain randomization, enabling the network to experience a wide spectrum of intensity profiles and imaging conditions.

The results of our evaluation on three publicly available and two in-house datasets shows that TimeLesSeg outperforms the contrast-agnostic state of the art on single-modality inputs across overlap- and distance-based metrics. In longitudinal processing, our method outperforms SAMSEG, and captures lesion load dynamics more accurately than both the former and LST-AI. All source code related to the development of our method is available at https://github.com/NeuroADaS-Lab/TimeLesSeg.

# 1. Introduction

Multiple Sclerosis (MS) is a heterogeneous immune-mediated inflammatory disease affecting the central nervous system. In MS, demyelinating plaques visualized in vivo using magnetic resonance imaging (MRI) represent the principal imaging biomarker of disease activity. Therefore, their precise detection and consistent longitudinal monitoring are fundamental for accurate diagnosis and effective disease management.

Given the subjective, monotonous, and labor-intensive nature of manual annotation, significant efforts have been dedicated in the last two decades to developing reliable automatic MS lesion segmentation tools (Lladó et al. 2012). Since the second half of the 2010s, deep learning-based techniques have dominated the state of the art, with convolutional neural networks (CNNs) being the architectural standard (Brosch et al. 2016; Havaei et al. 2016). Currently, however, they possess significant limitations that restrict their applicability in clinical settings.

First, current deep learning-based methods lack the flexibility to adapt to the distinct segmentation contexts required in MS. This limitation stems from the dual role MRI plays in managing the disease: in *cross-sectional* analysis (diagnostic setting), a single-timepoint snapshot of a subject is segmented; whereas in *longitudinal* analysis, (monitoring) spatiotemporal correlations across multiple single-subject MRI acquisitions are exploited to identify new radiological activity (Diaz-Hurtado et al. 2022). The state of the art reflects this dichotomy, as most methods are designed to operate in either cross-sectional or longitudinal settings, but not both.

Second, deep learning models tend to learn highly domain-specific features. The inherent MRI heterogeneity, combined with the difficulties in curating large, diverse datasets, renders methods particularly susceptible to covariate shifts. Consequently, they struggle to generalize to out-of-distribution data, even within the same imaging modality (Valverde et al. 2019). Collectively, this structural rigidity and lack of generalizability restrict their translation to clinical practice and research, particularly in longitudinal studies, where issues such as mid-study distribution shifts or missing data – whether modalities or timepoints – can render a method unreliable or even inoperable.

Various approaches have been developed to improve robustness to changes in contrast. A domain randomization-based method demonstrated that a CNN trained on synthetic, nonrealistic MRI volumes can yield state-of-the-art segmentation performance of normal-appearing brain tissues, regardless of contrast and resolution (Billot et al. 2023). This approach was extended to also detect white matter hyperintensities (WMHSynthSeg), with promising results (Laso et al. 2023). Alternatively, SAMSEG (Puonti et al. 2016) achieves sequence adaptiveness by decoupling the modeling of neuroanatomical structure shape from their contrast-specific appearance on MRI. It was also extended to segment MS lesions through the incorporation of a lesion shape prior specified by a variational autoencoder (Cerri et al. 2021). More recently, SAMSEG was updated to perform longitudinal segmentation without constraints on the number of timepoints through the construction of a subject-specific atlas, which serves as a spatial constraint to ensure anatomical consistency (Cerri et al. 2023).

In contrast, domain-specific methods impose severe constraints on input dimensionality and type, i.e., training and test domains must remain as close as possible. The most common type of input is a single-timepoint combination of fluid-attenuated inversion recovery (FLAIR) and T1-weighted, as exemplified by *nicMSlesions (Valverde et al. 2017)* and LST-AI (Wiltgen et al. 2024).

Moving beyond traditional approaches to MS lesion segmentation, in this work we propose a novel formulation of the task that improves generalization by replacing MRI-based temporal encoding with a pathological prior derived from lesion masks. This represents a conceptual shift that leads to three key contributions. First, it enables joint cross-sectional and longitudinal lesion segmentation with the same method, operating seamlessly out-of-the-box without the need for domain adaptation or retraining. Second, by decoupling longitudinal modelling from the radiological appearance of lesions, our approach mitigates inconsistencies across timepoints, such as variations in scanner, imaging modality, and artifacts. Finally, the method simplifies the task of modeling lesion evolution patterns in the binary space, abstracting away their underlying modality-specific appearance.

Hence, in conjunction with our new segmentation framework, we devise a simple yet effective module that generates endless plausible prior stages by stochastically deforming a lesion mask. Inspired by a previously described generative method for training contrast-agnostic CNNs (Billot et al. 2023), we synthesize a sizable longitudinal pathological dataset starting with a set of nine brain segmentations, with which we train a single convolutional neural network that can segment any contrast and in both longitudinal and cross-sectional settings. We validate the resulting method on five different real datasets: ISBI (Carass et al. 2017), MSLESSEG (Guarnera et al. 2025), OPENMS (Lesjak et al. 2018), a clinical trial dataset (MS-SMART, Chataway et al. 2020), and a clinical cohort (Hospital Clínic de Barcelona) – demonstrating robustness to changes in domain and setting. We compare our work against other established methods, both contrast-specific and -adaptive, such as LST-AI, nicMSlesions, SAMSEG, and WMHSynthSeg.

# 2. Proposed method

Our framework, TimeLesSeg, through a straightforward, conceptual shift, generalizes the task of MS lesion segmentation across both temporal and contrast dimensions. The result is a single deep learning model able to robustly accommodate all the variability present in clinical and research MRI data, such as differing modalities, protocols, spatial resolution, and longitudinal scanning.

Our method integrates a pathological spatial prior in the form of a lesion mask, corresponding to the longitudinal scenario. Accordingly, cross-sectional segmentation – where no prior information is available – is represented using an empty lesion mask. This enables a joint framework for modeling cross-sectional and longitudinal segmentation with the same deep learning model. Moreover, generating longitudinal data from people with MS is significantly simplified in the binary space, as lesion trajectories are modeled through morphological operations applied to the lesion mask, simulating potential evolutions. We leverage this property in combination with a previously established MRI generation approach (Billot et al. 2023), allowing us to train a deep neural network on a fully-synthetic dataset generated from only nine unique segmentations. In the following sections, we outline this framework and provide a detailed account of the generative modules that constitute it.

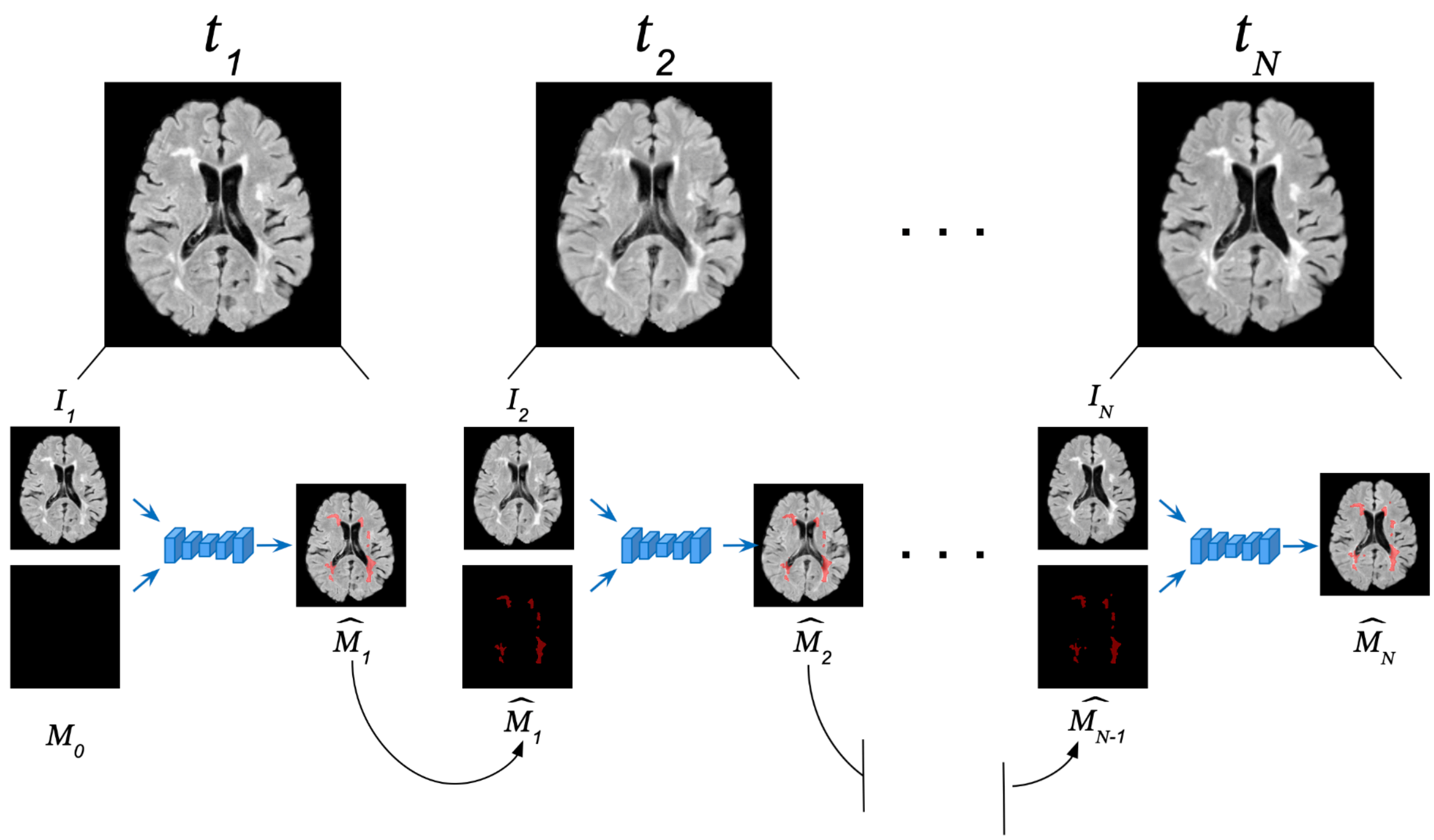

**Figure 1**. *Execution workflow for segmenting N timepoints with our proposed method, showcasing its flexibility. The first timepoint is processed cross-sectionally, as no prior information is available – note* $M_0$ *is an empty mask. Afterwards, the prediction is propagated in time. Note also that, although in the figure all MRI volumes belong to the same modality (FLAIR), this need not be the case.*

## 2.1. Jointly modeling cross-sectional and longitudinal scenarios

Both cross-sectional and longitudinal inputs are segmented with the same CNN as follows. The number of input channels of the network is set to two, the first of which contains the scan that represents the segmentation target, while the second one provides a disease-stage prior in the form of a lesion mask. Crucially, as previously mentioned, when no prior information is available, this second channel is set to an empty mask, in which case the CNN focuses only on the scan, reducing to a single-timepoint scenario.

More formally, let $f$ denote the follow-up, current, or target timepoint; and $b$ the baseline, representing any prior timepoint – i.e. $b \in \{0 \leq b' < f\}$. Let $I_f$ denote an image or scan at timepoint $f$, while $\mathcal{M}_b$ refers to a lesion mask at timepoint $b$. The input to our method is always a two-channel input, where the first channel corresponds to $I_f$ and the second one to $\mathcal{M}_b$. Crucially, the case where $b = 0$ (pre-diagnosis), is modeled by setting $\mathcal{M}_b = \emptyset$. Therefore, the segmentation model learns to derive $\mathcal{M}_f$ from $\{I_f, M_b\}$, where during training a fraction of all $M_b$ has been set to empty.

## 2.2. Longitudinal MS data generation via stochastically-applied morphological operations

To enlarge the space of lesion evolution patterns the CNN learns to segment, we devised a novel data augmentation module that stochastically deforms each individual lesion in a binary lesion segmentation, simulating plausible masks corresponding to prior timepoints. In accordance with the patterns observed in people with MS, four main categories are modeled: lesion enlarging, lesion shrinking, new lesion appearance, and stable lesion. These are respectively modeled through morphological operations: binary erosion, binary dilation, replacement by background, and identity transform. Notably, since augmentations are applied to simulate a prior timepoint (i.e., reverse temporal order), the morphological operations are purposefully the opposite of what intuition might suggest – e.g., lesion enlargement is simulated by erosion. We will hereinafter refer to this augmentation module as *Fake Lesion Mask*, or *FLM* for short.

In order to ensure that the network is exposed to realistic frequencies of each evolution pattern, augmentations were selected according to the lesion's volume – e.g., when producing plausible prior timepoints, lesions larger than 2500 $mm^3$ are always modeled as stable – as well as a set of predefined probabilistic cutoffs, which are compared against a random sample from a uniform distribution. The exact set of cutoffs that were used for the experiments described in this manuscript is given in Section 3.1.

## 2.3. Synthetic scan generation using Gaussian Mixture Models

To provide robustness to scanner-, site-, and protocol-induced variability, whilst improving generalization across heterogeneous clinical domains, we employ a generative domain-randomization based strategy (Billot et al. 2023). Its design philosophy is that, by training a network on a dataset with maximal variability, any real clinical domain (i.e., imaging modality) will be encompassed within the training distribution, thereby eliminating the possibility of a domain shift altogether.

In short, the input segmentation is first subjected to strong spatial augmentation using a combination of a random affine and a non-linear deformation. A synthetic image is then generated by sampling from a randomly parameterized Gaussian mixture model (GMM) conditioned on the augmented segmentation, followed by bias-field corruption to mimic intensity inhomogeneities. Finally, realistic acquisition variability is simulated through anisotropic resampling with random plane, slice spacing, and thickness, before restoring to the original resolution. Figure 2 displays an example scan obtained using this method.

## 2.4. Full generative pipeline

The different components described above – *FLM* and GMM – are integrated as follows (see Figure 2 for a visual representation). The input to the full generative method is a set of $N$ tuples, each containing a brain parcellation ($\mathcal{S}$) with $K$ classes and a binary lesion mask ($\mathcal{M}$): $\mathcal{L} = \{\mathcal{S}, \mathcal{M}\}$, where the value at their $j^{\text{th}}$ voxel verify $\mathcal{S}^j \in \{1, ..., K\}$ and $\mathcal{M}^j \in \{0, 1\}$.

First, we apply the *FLM* module $M$ times to each $\mathcal{M}$. At this point in the process, the goal is not to generate biologically plausible prior disease stages, but to increase the variability in lesion spatial configurations. Thus, the probabilities of each lesion evolution pattern are tailored to perform more aggressive augmentation.

Afterwards, we merge each augmented lesion mask $\widehat{\mathcal{M}}$ into its corresponding $\mathcal{S}$, preventing lesion assignment for those voxels in anatomically implausible regions (e.g., cerebrospinal fluid). This combined segmentation is subsequently used to generate $P$ synthetic scans by sampling from the randomly parameterized Gaussian mixture model (GMM) described in Section 2.3.

This yields a synthetic scan ($I_f$) and a spatially deformed lesion mask ($\mathcal{M}_f$). Finally, a more realistically-parameterized *FLM* is applied $L$ times to $\mathcal{M}_f$, obtaining $L$ plausible prior timepoints ($\mathcal{M}_b$). Once again, after each execution of *FLM*, we ensured that lesions had not encroached upon anatomically impossible regions. This yields, from the initial set of $N$ segmentations, $N \times M \times P \times L$ training $\{I_f, \mathcal{M}_b, \mathcal{M}_f\}$ triplets, where $\mathcal{M}_b$ is a lesion mask at a previous timepoint, and $\mathcal{M}_f$ the ground truth for image $I_f$, both at follow-up.

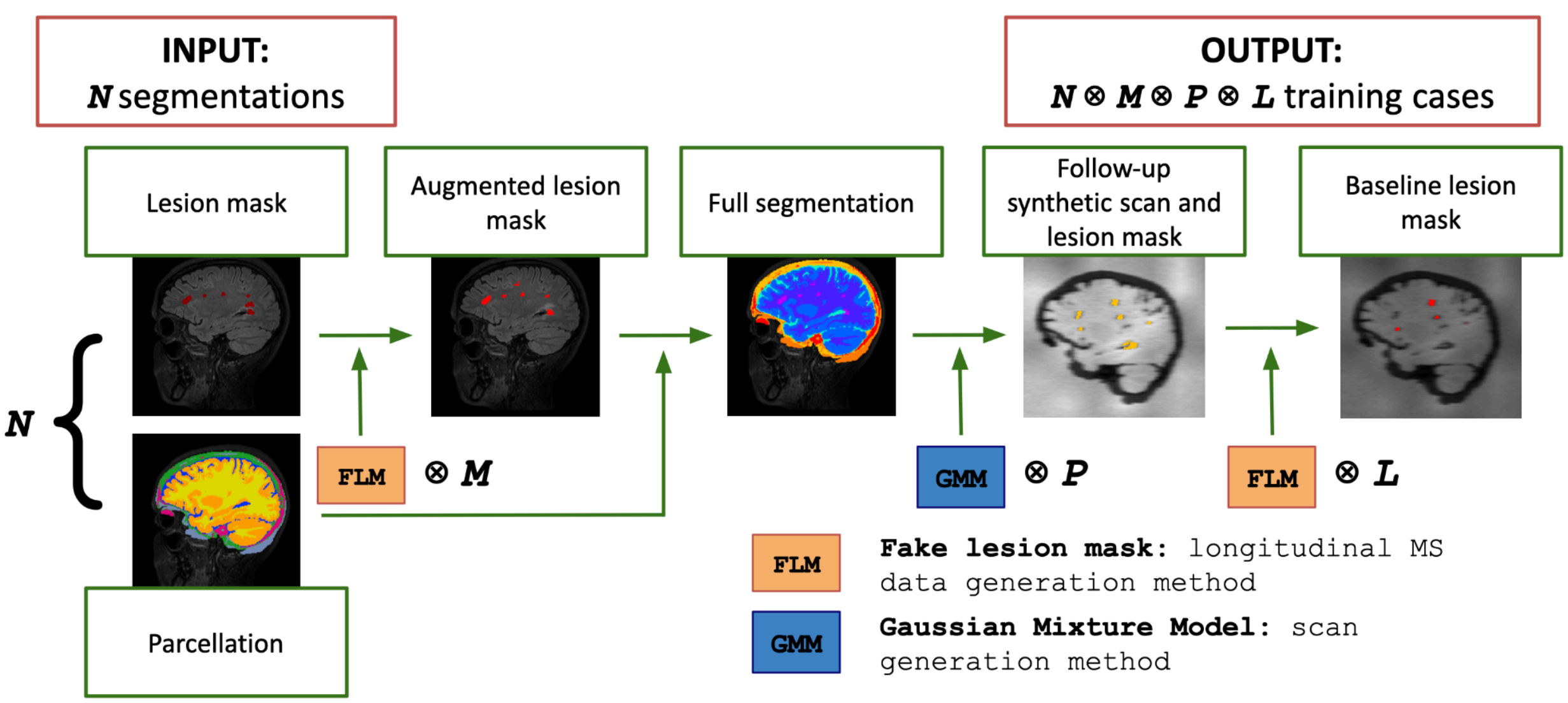

**Figure 2**. *Generative pipeline used to synthesize a longitudinal dataset with synthetic MRI scans from people with MS following our framework. The required input is a set of segmentations, each containing a brain parcellation and an MS lesion mask. First, the lesion mask is augmented with Fake Lesion Mask (FLM) with parameters adjusted to augment the space of lesion loads and configurations that will appear in the synthetic scans. Then, the resulting lesion mask and its corresponding parcellation are joined together and fed to the domain randomization-based scan generation method, which spatially deforms them prior to sampling from a Gaussian Mixture Model (GMM) conditioned on it. This yields a tuple of scan and lesion mask – both understood to be at follow-up, the latter of which is augmented with FLM generating a prior timepoint – the baseline lesion mask.*

# 3. Experimental setup

## 3.1. Implementation details

### 3.1.1 Synthetic Dataset Generation

Starting with a set of nine semi-automatic segmentations, each with a brain parcellation and a lesion mask, we generated 16875 $\{I_f, \mathcal{M}_b, \mathcal{M}_f\}$ training triplets by setting, following Section 2.4's notation, $M = 15$, $P = 25$, and $L = 5$.

The lesion masks had been manually annotated, while the brain parcellations had been derived using Geodesic Information Flow (Cardoso et al. 2015) following the Desikan-Killiany-Tourville atlas (Klein and Tourville 2012). The classes used were: gray matter, white matter, cerebrospinal fluid, thalamus, ventral diencephalon, brain stem and pons, three classes with non-brain tissues, and white matter lesions.

Baseline masks are simulated through morphological operations applied to each individual lesion in a lesion mask. The probability of simulating a given lesion evolution pattern depends not only on a set of predefined probabilistic cutoffs, but also on the input lesion's volume. For the first run of *FLM*, the cutoffs were defined as follows. First, 40% of lesions were modeled as stable regardless of their size. From the remaining 60%, probabilities were assigned based on volume categories. Lesions with a volume less than $200$ mm$^3$ ($v < 200$ mm$^3$) were modeled as stable with an additional probability of $0.5$ (total $P = 0.9$), and simulated as new lesions otherwise ($P = 0.1$). Medium-sized lesions ($200 \leq v < 1000$ mm$^3$) were modeled as stable with an additional probability of 0.3 (total $P = 0.7$), or either eroded for 1 iteration, dilated for 1 iteration, or simulated as a new lesion, each with $P = 0.1$. Finally, larger lesions ($v \geq 1000$ mm$^3$) were simulated as stable with an additional probability of $0.15$ (total $P = 0.55$), or either eroded for 1 voxel, dilated for 1 voxel, or modeled as new lesion, all with $P = 0.15$.

The realistic hyperparameters invert the previous volume-dependent logic. Whereas before smaller lesions were preferentially modeled as stable, this configuration prioritizes stability for very large lesions ($v > 2500$ mm$^3$), reflecting the observation that large, confluent lesions rarely exhibit rapid morphological changes. As before, we allocate a baseline volume-independent probability of stability, this time of 30%. Lesion growth (simulated with erosion) is modeled as the predominant pattern in MS lesions – with a total probability $0.45$. Specifically, single-voxel erosion is performed with $P = 0.35$, two-voxel erosion with $P = 0.08$, and three-voxel erosion with $P = 0.02$. For small lesions ($v < 250$ mm$^3$), we cap the number of erosion iterations to two to avoid overrepresenting new lesion appearance, which is modeled separately with $P = 0.24$. Finally, as lesion shrinking is the least likely pattern in people with MS, binary dilation is performed with $P = 0.01$.

Synthetic domain-randomized scans were generated from a GMM conditioned on a spatially augmented label map, both randomly parameterized. The hyperparameters governing these processes are sampled from uniform distributions, the bounds of which are set as follows. Spatial deformations: rotation (degrees) $\sim \mathcal{U}(-15, 15)$, scaling (mm) $\sim \mathcal{U}(0.8, 1.2)$, shearing (mm) $\sim \mathcal{U}(-0.012, 0.012)$, nonlinear SVF standard deviation $\sim (0, 4)$. GMM parameters: $\mu_k \sim \mathcal{U}(0, 250)$, $\sigma_k \sim \mathcal{U}(0, 30)$. Resolution randomization parameters: probability anisotropic acquisition simulation $= 0.9$, maximum anisotropy spacing = $5$ mm. Scans were clipped to a maximum intensity of $300$ after generating them.

We ensured that lesions were discernable from white matter (WM) using an approach inspired on statistical measures such as the t-test and Cohen's $d$ for measuring effect sizes. Namely, once a scan had been generated, we computed the difference of means divided by the sum of standard deviations between intensities in WM and lesion voxels, namely:

$$EF = \frac{|\mu_{WM} - \mu_l|}{\sigma_{WM} + \sigma_l}$$

The scan was accepted as valid if and only if $EF$ between lesional tissue and WM was greater than the 80$^{th}$ percentile of all pairwise EFs between all classes in the parcellation.

### 3.1.2 Model Architecture

We employed a UNet with five resolution stages and residual connections in both encoder and decoder. The initial number of features is set to 32, and doubled at each stage except for the deepest or lowest resolution stage, which is capped to 320, mimicking nnUNet's logic (Isensee et al. 2021).

Each resolution stage consists of two ResNetD residual blocks (He et al. 2018), with each block defined as the sequential application of two Conv → Norm → ReLU layers and a residual connection. Downsampling in the encoder is implemented with stride two convolutions, while upsampling in the decoder is performed with transposed convolutions. In case of mismatch in the number of features or input dimensions between the block's output and the residual connection, the latter is projected, respectively, with a kernel 1 convolution or average pooling operation with kernel size 2 and stride 2. The input patch size of the network is set to [128, 128, 96] voxels.

### 3.1.3 Model training

The objective function used to train our network was the sum of soft Dice and binary cross entropy losses. We applied deep supervision exclusively at the first two resolution stages, as further downsampling caused small lesions to vanish. The learning rate was initialized at $\eta = 0.01$ and decayed using the *poly* scheduler (Isensee et al. 2021), which is defined at epoch $e$ as $\eta_e = \eta \times (1 - \frac{e}{e_T})^{0.9}$, where $e_T = 2000$ is the number of epochs. Optimization was conducted using Stochastic Gradient Descent with 0.99 Nesterov momentum and a $l_2$ weight decay of $3e^{-5}$. A training epoch was defined as 250 iterations, while a validation epoch consisted of 50.

### 3.1.4 Inference

Our method requires minimal preprocessing thanks to the high heterogeneity in its synthetic training dataset. Namely, images are reoriented closest to RAS world space, Z-score intensity normalized, and then resampled with bicubic interpolation to an isotropic [1., 1., 1.] mm spacing. Afterwards, a prediction is obtained using a sliding window approach with a step size of 0.5, gaussian blending of logits, and mirroring as a test-time augmentation strategy. When multiple modalities are provided to the method, their predicted probabilities are averaged with user-specified weights – but by default, in this work we use equal weights.

## 3.2. Validation datasets

To provide a fair and standardized method comparison, the methods described in Section 3.3 were evaluated on five datasets. Three of them are publicly available, while the other two are from clinical sources and are thus subjected to access restrictions. One was acquired at Hospital Clínic, Barcelona, Catalunya, Spain; while the other one is from a clinical trial conducted in multiple sites across the UK.

### 3.2.1. Hospital Clinic

Comprises data from 117 people with MS, each with two timepoints and two modalities: 3D FLAIR and 3D T1-weighted at 1 $mm^3$ resolution. Images were acquired at Hospital Clínic, Barcelona, Catalunya, Spain. Ground truth segmentations for each timepoint were generated with LST-AI as a starting point and subsequently refined by human experts.

### 3.2.2. ISBI

Contains scans from five people with MS, all acquired with a 3T Phillips scanner. Four subjects have data from four timepoints, and the other one five. The time between consecutive timepoints was approximately one year for all subjects. Each subject has four modalities per timepoint: fluid attenuated inversion recovery (FLAIR), T1-weighted, proton density (PD) and T2-weighted. All images have been bias field corrected (N4 algorithm), skull-stripped, dura-stripped, and registered to the 1$mm^3$ MNI template (Carass et al. 2017).

### 3.2.3. MSLESSEG

Includes imaging data from 75 people with MS, with ages ranging from 18 to 59 (Guarnera et al. 2025), and from different sources with scanners at 1.5 and 3T. Out of the 75 total subjects, 50 have a single timepoint, 15 have two, five have three, and five have four timepoints. Each timepoint includes three modalities: FLAIR, T2-weighted and T1-weighted. All data have been registered to the 1$mm^3$ MNI152 template using FLIRT and brain extracted with BET (Jenkinson et al. 2002; Smith 2002). Manual annotations were performed primarily on FLAIR.

### 3.2.4 OPEN-MS

Openly available dataset hosted on the website of the Laboratory of Imaging Technologies of the Faculty of Electrical Engineering, University of Ljubljana, Slovenia (https://lit.fe.uni-lj.si/en/research/resources). It comprises data from a cohort of 30 people with MS, acquired using a 3T Siemens scanner at the University Medical Center Ljubljana (Lesjak et al. 2018). Each subject has 3 modalities: 2D T1-weighted, 2D T2-weighted and 3D FLAIR scans. Preprocessing consisted of intra-subject affine registration of T1w and T2w scans to the FLAIR space using third-order interpolation. After registration, the authors performed bias field correction. Semi-manual segmentation was done by three raters on axial FLAIR slices, with both T1w and T2w axial slices displayed side-by-side.

### 3.2.5 MS-SMART

Private dataset from a phase 2b, multiarm, parallel group, double-blind, randomized placebo-controlled trial on the effects of three potentially neuroprotective drugs, with MRI data acquired across different clinical neuroscience centers in the UK in people with Secondary Progressive Multiple Sclerosis (SPMS) (Chataway et al. 2020) (ClinicalTrials.gov ID NCT01910259). Ground truth lesion masks were manually delineated on PD-weighted images and subsequently resampled to the corresponding FLAIR space.

## 3.3 Benchmark methods

### 3.3.1. LST-AI

Deep learning-based MS lesion segmentation tool that requires a two-channel input with a T1-weighted and a fluid attenuated inversion recovery (FLAIR) scan (Wiltgen et al. 2024). It employs an ensemble with three 3D UNets, each with 5 resolution stages. The final segmentation is obtained by averaging all three independently-produced probability arrays and then applying a user-controlled threshold, which by default is set to 0.5. Each network was trained separately with deep supervision and a combination of soft Dice and binary cross entropy losses. Preprocessing is handled by the tool, and consists of a registration to MNI152 1x1x1 mm space and skull-stripping with HD-BET. Optional lesion location annotation is provided using the MNI atlas brain parcellation.

### 3.3.2. SAMSEG

Sequence Adaptive Multimodal Segmentation (SAMSEG) is the only non-deep learning benchmark method (Puonti et al. 2016). Its main advantage, sequence adaptiveness, relies on not making any assumptions about the prior distribution of image intensity parameters. Instead, it segments in an unsupervised manner, optimizing simultaneously a fit of a deformable probabilistic tetrahedral atlas mesh (neuroanatomical shape prior), a per-class mixture of gaussian components, and a spatially smooth bias field modeled through a sum of cosine basis functions.

The base segmentation method was extended to model also white matter lesions (Cerri et al. 2021). A prior on lesion shape is introduced through a Variational Autoencoder, while lesion location probabilities are estimated from a training set in a procedure similar to the one used to derive the neuroanatomical segmentation. The final segmentation is then performed by estimating first the lesion segmentation, and then the normal tissue segmentation.

In our cross-sectional evaluation experiments (results reported in Sections 4.1 and 4.2), we used SAMSEG's cross-sectional model (command run_samseg), while for the longitudinal experiments (Section 4.3) the longitudinal version of the tool was used (run_samseg_long).

### 3.3.3. WMHSynthSeg

Extension of SynthSeg (Billot et al. 2023) that also segments white matter lesions (Laso et al. 2023). Following a domain randomization paradigm, the authors trained a 3D UNet for approximately $10^5$ iterations on synthetic data generated on-the-fly by sampling from a randomly-parameterized mixture of gaussian distributions conditioned on a segmentation. The authors ensured that areas with white matter hyperintensities were distinguishable from normal-appearing white matter on scans by imposing a threshold-based constraint on the values that the means of both tissues could take.

### 3.3.4. nicMSlesions

Cascaded ensemble of two relatively lightweight convolutional neural networks (~500k parameters), each with two stacks of convolutional and maxpooling layers, followed by a fully-connected layer that produces a two-dimensional output: the probability of the voxel at the center of the input 11x11x11 patch being either healthy or lesional tissue (Valverde et al. 2017). The first network provides an initial segmentation of all voxels, and those with a probability over 0.5 of belonging to lesional tissue are then processed by the second CNN, which has been trained specifically to refine false positives. The version we tested in this paper uses a single-modality FLAIR input.

## 3.4 Validation metrics

The performance of all automated algorithms considered in this manuscript was done with counting/overlap-based metrics as well as distance-based metrics in accordance with medical image segmentation guidelines (Commowick et al. 2018; Maier-Hein et al. 2024). Furthermore, since voxel-level Dice Similarity Coefficient (DSC) falls short when evaluating an algorithm's ability in detecting lesions as separate individual entities, we also compute lesion-wise DSC. Only lesions with a volume over 3mm$^3$ were considered for our evaluation.

### 3.4.1 Overlap measures

Given a ground truth $\mathcal{G}$ and a method's prediction $\mathcal{P}$, overlap or *counting*-based metrics capture the number of entities both agree on. Let | . | refer to a set's cardinality, and $TP$, $FP$, $FN$, and $TN$ to the total number of true positives, false positives, false negatives, and true negatives, respectively. Positive Predictive Value (PPV) measures the probability of a voxel belonging to a lesion given a method's classification as such, and is calculated as the rate of positively predicted voxels which are actually positive, i.e.,

$$PPV = \frac{|\mathcal{G} \cap \mathcal{P}|}{|\mathcal{P}|} = \frac{TP}{TP+FP}$$

False Positive Rate (FPR), on the other hand, quantifies the number of false positives a method commits, divided by the total number of negative events (healthy voxels), i.e.,

$$FPR = \frac{|\mathcal{G}^c \cap \mathcal{P}|}{|\mathcal{G}^c|} = \frac{FP}{TN+FP}$$

To overcome the last two metrics' susceptibility to class imbalance, Dice Similarity Coefficient (DSC) employs the harmonic mean of *PPV* and 1 - *FPR* (Sensitivity), thus penalizing extreme values of either metric. DSC is defined as follows:

$$DSC = \frac{2 \times |\mathcal{G} \cap \mathcal{P}|}{|\mathcal{G}| + |\mathcal{P}|} = \frac{2 \times TP}{2 \times TP + FP + FN}$$

Since object (lesion) detection is arguably more clinically relevant than voxel-wise overlap, we also quantify method performance with Lesional DSC. This metric is computed with the same formula as DSC, only differing in how $TP$, $FP$, and $FN$ are defined. It requires that both ground truth ($\mathcal{G}$) and prediction ($\mathcal{P}$) contain uniquely labeled connected components. Then, true positives are defined as the number of connected components in $\mathcal{G}$ that have at least a one voxel overlap with any nonzero voxel in $\mathcal{P}$. On the other hand, false positives are calculated as those labels in $\mathcal{P}$ that have no overlapping foreground voxel in $\mathcal{G}$. Finally, false negatives are those identifiers in $\mathcal{G}$ that have no overlapping foreground voxel in $\mathcal{P}$ (Gaj et al. 2021).

### 3.4.3 Distance measures

To provide a more precise estimation of distances between two surfaces, we use DeepMind's approximation (Nikolov et al. 2021). Let $\mathcal{S}_\mathcal{G}$ and $\mathcal{S}_\mathcal{P}$ denote the surface representations (sets of surfels) of $\mathcal{G}$ and $\mathcal{P}$, respectively, where $\mathcal{S}_\mathcal{G}$ is comprised of $N$ surfels, while $\mathcal{S}_\mathcal{P}$ has $M$ surfels. Each surfel is described by a tuple containing both its spatial position $\mathbf{x} \in \mathbb{R}^3$ and its surface area $a \in \mathbb{R}^+$. Let $(\boldsymbol{g}_i, \alpha_i)$ denote the spatial position and surface area of the $i^{th}$ surfel in $\mathcal{S}_\mathcal{G}$, and $(\boldsymbol{p}_j, \beta_j)$ be the corresponding tuple describing the $j^{th}$ surfel in $\mathcal{S}_\mathcal{P}$. Then, the distance between the $i^{th}$ surfel in $\mathcal{S}_\mathcal{G}$ and $\mathcal{S}_\mathcal{P}$ is given by:

$$d(\boldsymbol{g}_i, \mathcal{S}_\mathcal{P}) = \min_{j \in \{1, \ldots, M\}} ||\boldsymbol{g}_i - \boldsymbol{p}_j||_2$$

We aggregate the sets of distances between two surfaces in two ways: using the area-weighted 95$^{th}$ percentile Hausdorff Distance (HD95) and the Average Symmetric Surface Distance (ASSD). HD95 is defined as follows. Let $F(\zeta)$ represent the fraction of the total surface area in $\mathcal{S}_\mathcal{G}$ that is within a distance $\zeta$ of $\mathcal{S}_\mathcal{P}$:

$$F(\zeta) = \frac{\sum_{i=1}^{N} \alpha_i \mathbb{I}(d(\boldsymbol{g}_i, \mathcal{S}_\mathcal{P}) \leq \zeta)}{\sum_{i=1}^{N} \alpha_i}$$

where $\mathbb{I}(\cdot)$ is the indicator function. The asymmetric $HD_{95}$ is then defined as follows:

$$HD_{95}(\mathcal{S}_\mathcal{G}, \mathcal{S}_\mathcal{P}) = \inf\{\zeta \mid F(\zeta) \geq 0.95\}$$

where $\inf$ refers to the infimum. The final symmetric $HD_{95}$ is obtained as the maximum between the two asymmetric ones: $\max(HD_{95}(\mathcal{S}_\mathcal{G}, \mathcal{S}_\mathcal{P}), HD_{95}(\mathcal{S}_\mathcal{P}, \mathcal{S}_\mathcal{G}))$.

Secondly, ASSD is computed by weighting the distances by their *surfel* areas and dividing by the total area in both surfaces.

$$ASSD(\mathcal{S}_\mathcal{G}, \mathcal{S}_\mathcal{P}) = \frac{\sum_{i=1}^{N} \alpha_i d(\boldsymbol{g}_i, \mathcal{S}_\mathcal{P}) + \sum_{j=1}^{M} \beta_j d(\boldsymbol{p}_j, \mathcal{S}_\mathcal{G})}{\sum_{i=1}^{N} \alpha_i + \sum_{j=1}^{M} \beta_j}$$

## 3.5 Statistical analysis

Differences in method performance is assessed using Wilcoxon Rank Sum's test, with a null hypothesis that the distributions of two methods differ by a location shift of 0. Multiple comparison correction was performed using Benjamini and Hochberg's method (also known as BH or FDR). Four statistical significance levels were defined and used throughout the Results section: *ns*, *, ** , and *** each corresponding to non-significant, $p < 0.05$, $p < 0.01$, and $p < 0.001$ differences, respectively. All analysis and figures were generated using R Statistical Software (v4.5.2; R Core Team 2026).

# 4. Results

The performance of TimeLesSeg was evaluated against the state of the art in MS lesion segmentation, using overlap-based metrics – on a voxel (Dice Similarity Coefficient, DSC), and object level (Lesional DSC) –, counting metrics – Positive Predictive Value (PPV) and False Positive Rate (FPR) –, and distance-based metrics – 95$^{th}$ percentile Hausdorff Distance (HD95) and Average Symmetric Surface Distance (ASSD). When a method predicted a null lesion load (0 mm$^3$, no lesions) for a given test case, we omitted it when aggregating distance-based metrics (since in that case the metric is undefined). However, for both overlap and counting metrics (i.e., DSC, lesional DSC, PPV, and FPR), a result of 0 was assigned. For subsections 4.3-4.4, where figures are used, results are reported as median ± interquartile range (IQR) throughout the section's text. Statistical significance was assessed with the Wilcoxon Rank Sum (also referred to as the Mann-Whitney) test for two non-parametric samples.

## 4.1 Single modality

On Table 1, the results for all contrast-agnostic methods on single-modality FLAIR inputs are shown per dataset and across all metrics. Our method consistently outperforms both WMHSynthSeg and SAMSEG on single-modality inputs on all metrics, except for PPV and FPR.

***Table 1.*** *Single-modality (FLAIR) results across Hospital Clínic, ISBI, MSLESSEG and OPENMS datasets, obtained with WMHSynthSeg, SAMSEG, and our proposed method. Results are shown as* median (IQR)*, where IQR stands for Interquartile Range. *, ** and *** indicate statistical significance with p < 0.05, p < 0.01 and p < 0.001 levels of significance, respectively, between the performance of each benchmark method versus ours. DSC: Dice Similarity Coefficient, PPV: Positive Predictive Value, FPR: False Positive Rate, HD-95: 95$^{th}$ percentile Hausdorff Distance, ASSD: Average Symmetric Surface Distance.*

| | DSC | Lesional DSC | PPV | FPR | HD-95 (mm) | ASSD (mm) |
|---|---|---|---|---|---|---|
| **Hospital Clínic** | | | | | | |
| WMHSynthSeg | 0.37 (0.23) *** | 0.10 (0.33) *** | 0.43 (0.35) *** | 4.91e-4 (3.85e-4) | 29.17 (13.17) *** | 5.00 (3.09) *** |
| SAMSEG | 0.13 (0.47) *** | 0.40 (0.23) *** | 0.49 (0.82) | 2.23e-6 (5.05e-5) *** | 37.00 (22.56) *** | 7.69 (8.18) *** |

| TimeLesSeg | 0.64 (0.26) | 0.69 (0.16) | 0.53 (0.29) | 4.54e-4 (4.14e-4) | 13.85 (19.33) | 2.28 (2.74) |
|---|---|---|---|---|---|---|
| **ISBI** | | | | | | |
| WMHSynthSeg | 0.39 (0.08) *** | 0.44 (0.15) ** | 0.56 (0.18) | 2.59e-4 (2.08e-4) | 33.20 (8.62) *** | 5.78 (0.76) *** |
| SAMSEG | 0.56 (0.17) | 0.38 (0.29) ** | 0.94 (0.10) *** | 2.18e-5 (4.04e-5) *** | 23.35 (20.24) *** | 3.56 (6.69) ** |
| TimeLesSeg | 0.66 (0.13) | 0.58 (0.28) | 0.72 (0.26) | 2.98e-4 (2.45e-4) | 13.93 (12.28) | 1.82 (1.32) |
| **MSLESSEG** | | | | | | |
| WMHSynthSeg | 0.29 (0.26) *** | 0.50 (0.16) *** | 0.23 (0.31) *** | 1.06e-3 (7.10e-4) *** | 36.73 (18.52) *** | 6.94 (6.71) *** |
| SAMSEG | 0.57 (0.35) | 0.53 (0.25) *** | 0.67 (0.27) * | 2.24e-4 (6.21e-4) *** | 27.59 (22.17) | 3.68 (6.76) |
| TimeLesSeg | 0.61 (0.20) | 0.67 (0.13) | 0.60 (0.28) | 4.12e-4 (6.50e-4) | 24.35 (18.14) | 3.29 (3.90) |
| **OPENMS** | | | | | | |
| WMHSynthSeg | 0.33 (0.21) *** | 0.29 (0.20) *** | 0.46 (0.44) *** | 5.16e-4 (6.01e-4) *** | 32.33 (6.72) *** | 6.75 (3.06) *** |
| SAMSEG | 0.49 (0.55) * | 0.17 (0.31) *** | 0.92 (0.25) * | 9.84e-6 (7.88e-5) *** | 24.34 (16.04) ** | 4.01 (10.95) * |
| TimeLesSeg | 0.61 (0.32) | 0.50 (0.13) | 0.81 (0.30) | 1.87e-4 (1.80e-4) | 18.81 (11.04) | 2.68 (3.05) |

## 4.2 Multimodal inputs (T1w and FLAIR)

Table 2 reports the performance from WMHSynthSeg, SAMSEG, LST-AI and TimeLesSeg on a combination of T1-weighted and FLAIR modalities. While LST-AI, the only contrast-specific method (i.e., it was trained on those very same two modalities), and SAMSEG predict simultaneously both contrasts, for WMHSynthSeg and our method each modality was segmented independently, and the resulting probabilities were averaged.

***Table 2.*** *Multi-modality (T1w + FLAIR) results across Hospital Clínic, ISBI, MSLESSEG and OPENMS datasets obtained with WMHSynthSeg, SAMSEG, LST-AI, and our proposed method. Results are shown as* median (IQR)*, where IQR stands for Interquartile Range. *, ** and *** indicate statistical significance with $p < 0.05$, $p < 0.01$ and $p < 0.001$ levels of significance, respectively, between the performance of each benchmark method versus ours. DSC: Dice Similarity Coefficient, PPV: Positive Predictive Value, FPR: False Positive Rate, HD-95: 95$^{th}$ percentile Hausdorff Distance, ASSD: Average Symmetric Surface Distance.*

| | Voxel-wise DSC | Lesion-wise DSC | PPV | FPR | HD-95 (mm) | ASSD (mm) |
|---|---|---|---|---|---|---|
| **Hospital Clínic** | | | | | | |
| WMHSynthSeg | 0.48 (0.24) *** | 0.39 (0.22) *** | 0.61 (0.38) *** | 1.66e-4 (1.23e-4) *** | 29.70 (13.80) *** | 5.05 (2.69) *** |
| SAMSEG | 0.63 (0.25) *** | 0.52 (0.21) *** | 0.64 (0.29) *** | 2.35e-4 (2.22e-4) *** | 15.78 (10.74) | 2.26 (2.70) *** |
| LST-AI | 0.74 (0.18) *** | 0.75 (0.18) *** | 0.73 (0.20) | 1.45e-4 (3.30e-4) | 8.86 (12.13) *** | 1.03 (1.47) *** |
| TimeLesSeg | 0.68 (0.20) | 0.61 (0.20) | 0.75 (0.22) | 1.30e-4 (1.85e-4) | 14.37 (15.43) | 1.79 (2.45) |
| **ISBI** | | | | | | |
| WMHSynthSeg | 0.32 (0.16) *** | 0.40 (0.07) *** | 0.68 (0.15) | 1.06e-4 (5.41e-5) | 38.76 (6.89) *** | 7.17 (3.04) *** |
| SAMSEG | 0.67 (0.15) ** | 0.55 (0.23) | 0.88 (0.07) ** | 1.04e-4 (1.70e-4) * | 14.63 (8.42) | 1.75 (1.60) |
| LST-AI | 0.66 (0.17) ** | 0.70 (0.23) | 0.90 (0.06) *** | 1.03e-4 (1.58e-4) * | 10.34 (15.28) | 1.72 (1.89) |
| TimeLesSeg | 0.54 (0.22) | 0.59 (0.29) | 0.73 (0.16) | 2.13e-4 (1.37e-4) | 15.30 (13.25) | 2.58 (3.08) |
| **MSLESSEG** | | | | | | |
| WMHSynthSeg | 0.30 (0.29) *** | 0.40 (0.21) *** | 0.39 (0.43) *** | 3.00e-4 (3.13e-4) *** | 40.03 (19.08) ** | 8.45 (6.91) ** |
| SAMSEG | 0.58 (0.31) *** | 0.54 (0.21) | 0.63 (0.31) *** | 3.48e-4 (7.41e-4) *** | 26.15 (18.60) * | 3.82 (4.74) * |
| LST-AI | 0.65 (0.26) *** | 0.67 (0.23) *** | 0.69 (0.19) * | 1.88e-4 (5.65e-4) *** | 18.15 (25.46) *** | 2.35 (4.61) *** |
| TimeLesSeg | 0.49 (0.31) | 0.56 (0.24) | 0.73 (0.19) | 9.17e-5 (2.55e-4) | 32.55 (20.85) | 5.05 (8.09) |
| **OPENMS** | | | | | | |
| WMHSynthSeg | 0.34 (0.21) * | 0.20 (0.15) | 0.63 (0.48) *** | 2.19e-4 (1.95e-4) *** | 37.48 (7.88) *** | 8.08 (4.59) *** |
| SAMSEG | 0.60 (0.35) | 0.22 (0.25) | 0.84 (0.22) | 1.51e-4 (2.58e-4) | 20.26 (11.29) | 2.80 (4.38) |
| LST-AI | 0.75 (0.17) *** | 0.70 (0.09) *** | 0.86 (0.09) | 1.73e-4 (2.70e-4) * | 9.05 (10.95) *** | 1.03 (1.95) *** |
| TimeLesSeg | 0.51 (0.31) | 0.27 (0.18) | 0.91 (0.18) | 4.32e-5 (1.04e-4) | 22.61 (13.07) | 4.90 (5.93) |

## 4.3 Longitudinal processing

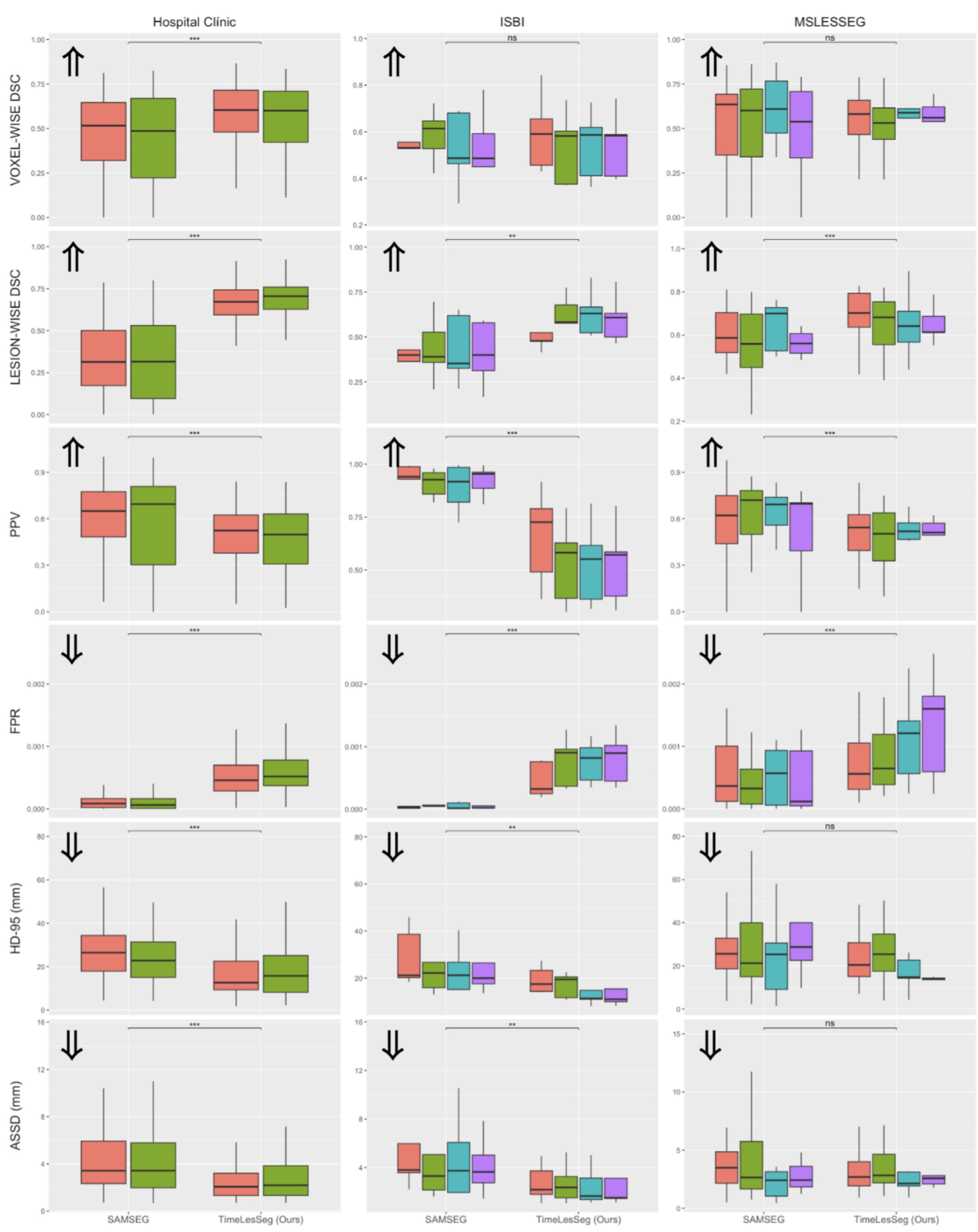

**Figure 3**. *Voxel-wise Dice Similarity Coefficient (DSC), Lesion-wise DSC, Positive Predictive Value (PPV), False Positive Rate (FPR), 95$^{th}$ percentile Hausdorff Distance (HD-95) and Average Symmetric*

*Surface Distance (ASSD) results for longitudinal processing methods (SAMSEG and TimeLesSeg) on Hospital Clínic de Barcelona and ISBI datasets (columns). Boxplots show the median (midline), the first and third quartiles (the hinges), and the furthest points ± 1.5 × IQR away from the hinges (whiskers). Each method's individual data points are overlaid. ns* indicates non-statistically significant differences, while **, ** and *** indicate statistical significance with $p < 0.05$, $p < 0.01$ and $p < 0.001$ levels of significance, respectively, between the performance of each benchmark method versus ours.*

We compare our method's longitudinal segmentation capabilities with the only other method with longitudinal processing capabilities: SAMSEG (Cerri et al. 2023). For TimeLesSeg, the results are obtained by segmenting the first timepoint cross-sectionally (with an empty mask), and then propagating the output segmentation sequentially in the next timepoints – exactly as depicted in Figure 1.

When we evaluated both methods on the dataset from Hospital Clínic (Figure 3, left-most column), SAMSEG scored a median Dice Similarity Coefficient (DSC) of 0.50 ± 0.40, while TimeLesSeg obtained 0.60 ± 0.27 ($p$ < 0.001). A similar trend emerges when contrasting both methods using Lesional DSC: SAMSEG, 0.31 ± 0.40; TimeLesSeg, 0.68 ± 0.15 ($p$ < 0.001). On Positive Predictive Value (PPV), SAMSEG achieved a median of 0.67 ± 0.37, while TimeLesSeg obtained 0.51 ± 0.29 ($p$ < 0.001). On the other hand, SAMSEG's False Positive Rate (FPR) (8.47e-5 ± 1.77e-4) was significantly lower than TimeLesSeg's (5.16e-4 ± 4.47e-4), with $p$ < 0.001. In terms of distance-based metrics, SAMSEG achieved a 95$^{th}$ percentile Hausdorff Distance (HD95) and an Average Symmetric Surface Distance (ASSD) of 25.45 ± 18.58 mm and 3.82 ± 4.80 mm, respectively; while TimeLesSeg achieved 14.90 ± 18.08 mm HD95 and 2.22 ± 2.74 mm ASSD ($p$ < 0.001 for both metrics).

On ISBI's public dataset (Figure 3, central column), comprising five subjects with over four timepoints each, SAMSEG obtained a median DSC of 0.53 ± 0.18, while TimeLesSeg scored 0.59 ± 0.21 (non-significant differences). However, on Lesional DSC TimeLesSeg did achieve a significantly higher score (0.58 ± 0.18 vs. 0.40 ± 0.27, $p$ < 0.01). On the other hand, on both PPV and FPR SAMSEG performed significantly better: 0.93 ± 0.12 and 4.68e-5 ± 4.68e-5, respectively; while TimeLesSeg obtained 0.58 ± 0.36 and 7.61e-4 ± 6.09e-4. On distance-based metrics, TimeLesSeg outperformed SAMSEG on both ASSD (1.79 ± 1.80 mm vs. 3.74 ± 3.86 mm, $p$ < 0.01) and HD95 (14.35 ± 9.45 mm vs. 21.19 ± 9.82 mm, $p$ < 0.01).

Finally, on dataset MSLESSEG (Figure 3, right-most column), SAMSEG's median DSC was 0.60 ± 0.36, while TimeLesSeg's was 0.57 ± 0.18 (*ns*). On Lesional DSC, SAMSEG achieved 0.57 ± 0.20 while TimeLesSeg achieved 0.68 ± 0.17 ($p$ 0.001). On both PPV and FPR, SAMSEG performed significantly better: 0.66 ± 0.31 vs TimeLesSeg's 0.52 ± 0.26 ($p$ < 0.001) and 3.65e-4 ± 7.73e-4 vs 6.46e-4 ± 8.62e-4 ($p$ < 0.001). On distance-based metrics, TimeLesSeg achieved 2.81 ± 3.10 mm, while SAMSEG 3.27 ± 4.51 mm (*ns*). When comparing HD95, both methods obtained statistically equivalent results: TimeLesSeg obtained a median of 19.42 ± 16.54 mm, and SAMSEG 25.02 ± 24.32 mm.

## 4.4 Clinical trial dataset

The generalizability of TimeLesSeg is further evaluated on a dataset belonging to a phase 2b, double-blind, randomized placebo clinical trial that aimed to assess the effects of amiloride, fluoxetine, and riluzole on neurodegeneration in people with secondary progressive multiple sclerosis (SPMS) (Chataway et al. 2020). We compare its performance with LST-AI and an in-house version of nicMSlesions (Valverde et al. 2017) maintained by University College London's Hawkes Institute.

In Figure 4, the results for 258 people with SPMS are shown, stratified per acquisition site. Non-significant differences in voxel-level Dice Similarity Coefficient were observed between LST-AI and TimeLesSeg, with respective median ± IQR of 0.58 ± 0.17 and 0.56 ± 0.26 (Figure 4, top-left quadrant). However, TimeLesSeg performed significantly better than nicMSlesions (0.48 ± 0.23, $p < 0.001$). In terms of Lesional DSC (Figure 3, second graph), LST-AI (0.65 ± 0.15) performed better than TimeLesSeg (0.62 ± 0.17, $p < 0.05$), which performed on par with nicMSlesions (0.64 ± 0.20).

In Positive Predictive Value (PPV), TimeLesSeg (0.49 ± 0.28) and LST-AI (0.48 ± 0.14) showed similar results, while nicMSlesions obtained a lower average (0.37 ± 0.22, $p < 0.001$ vs our method). When evaluating the rate of false positives each method committed, all three methods showed comparable results ($8.06e^{-4} \pm 8.53e^{-4}$ for TimeLesSeg; $1.01e^{-3} \pm 1.19e^{-3}$ for LST-AI; and $1.34e^{-3} \pm 1.27e^{-3}$ for nicMSlesions).

When comparing their performance on the 95$^{th}$ percentile Hausdorff's Distance (bottom-left quadrant), TimeLesSeg performed significantly worse than LST-AI ($p < 0.05$), the former with an average of 14.81 ± 13.38 mm while the latter scored 12.23 ± 12.93 mm. NicMSlesions scored an average HD95 of 14.54 ± 16.76 mm (*ns*). Finally, on ASSD our method achieved a significant ($p < 0.05$) lower results compared to nicMSlesions, 1.79 ± 2.12 mm vs 2.23 ± 2.62 mm, respectively. LST-AI's average was statistically equivalent to our method's average, with 1.51 ± 1.34 mm (Figure 4, bottom-right entry).

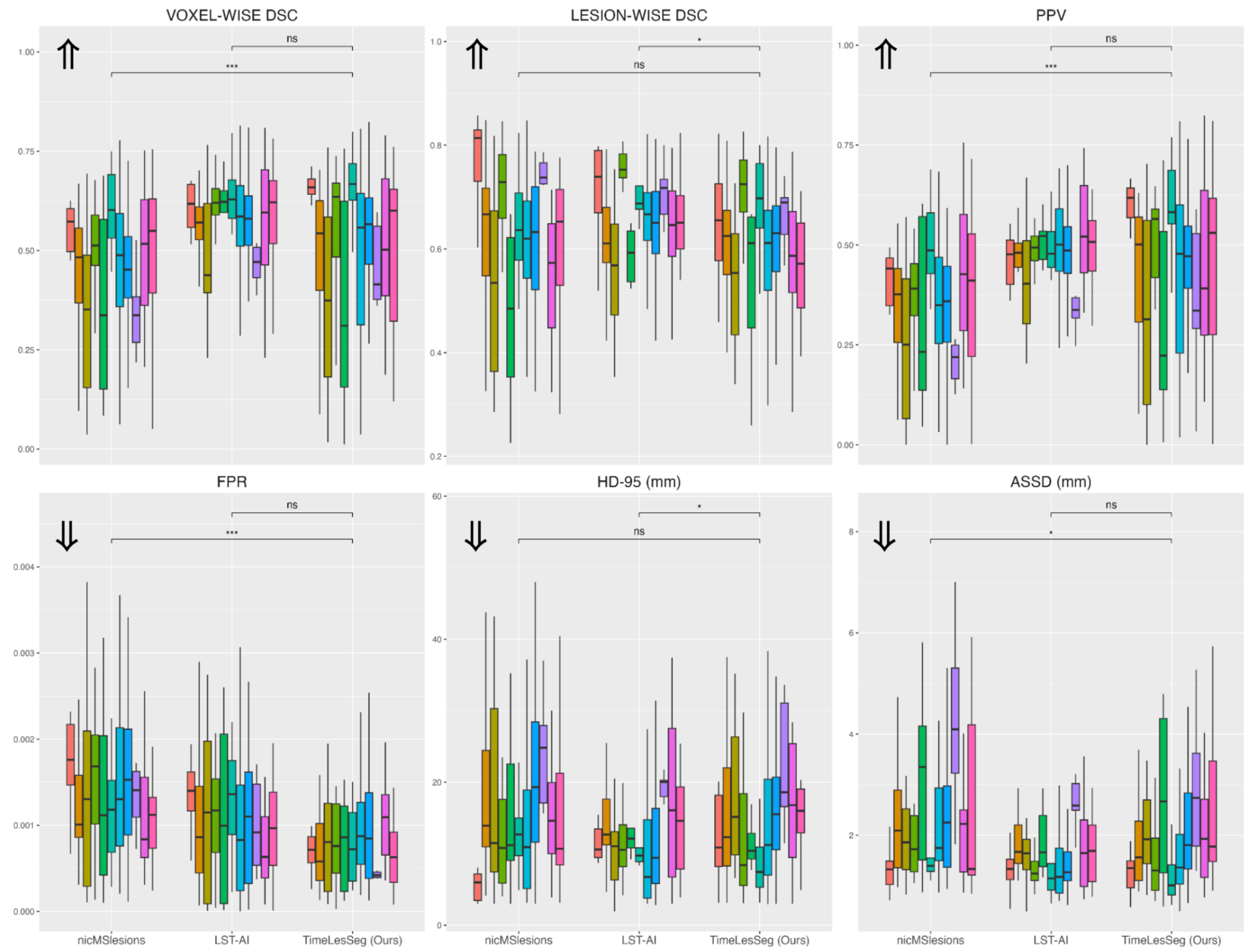

**Figure 4**. *From left to right, top to bottom: Voxel-wise Dice Similarity Coefficient (DSC), Lesion-wise DSC, Positive Predictive Value (PPV), False Positive Rate (FPR), 95$^{th}$ percentile Hausdorff Distance (HD-95), and Average Symmetric Surface Distance (ASSD) results on MRI scans from 258 people with secondary progressive MS from a clinical trial conducted in the UK (MS-SMART). Results are stratified per acquisition site or source, and correspond to T1w and FLAIR scans for LST-AI, and single-modality FLAIR scans for both nicMSlesions and TimeLesSeg. Boxplots show the median (midline), the first and third quartiles (the hinges), and the furthest points ± 1.5 × IQR away from the hinges (whiskers). Each method's individual data points are overlaid. ns* indicates non-statistically significant differences, while **, ** and *** indicate statistical significance with $p < 0.05$, $p < 0.01$ and $p < 0.001$ levels of significance, respectively, between the performance of each benchmark method versus ours.*

Using a Bland-Altman plot (Figure 5), TimeLesSeg demonstrates reduced bias and narrower limits of agreement compared to nicMSlesions and LST-AI, indicating improved volumetric consistency with the reference standard. All methods display heteroscedasticity, with increasing discrepancies at higher lesion volumes.

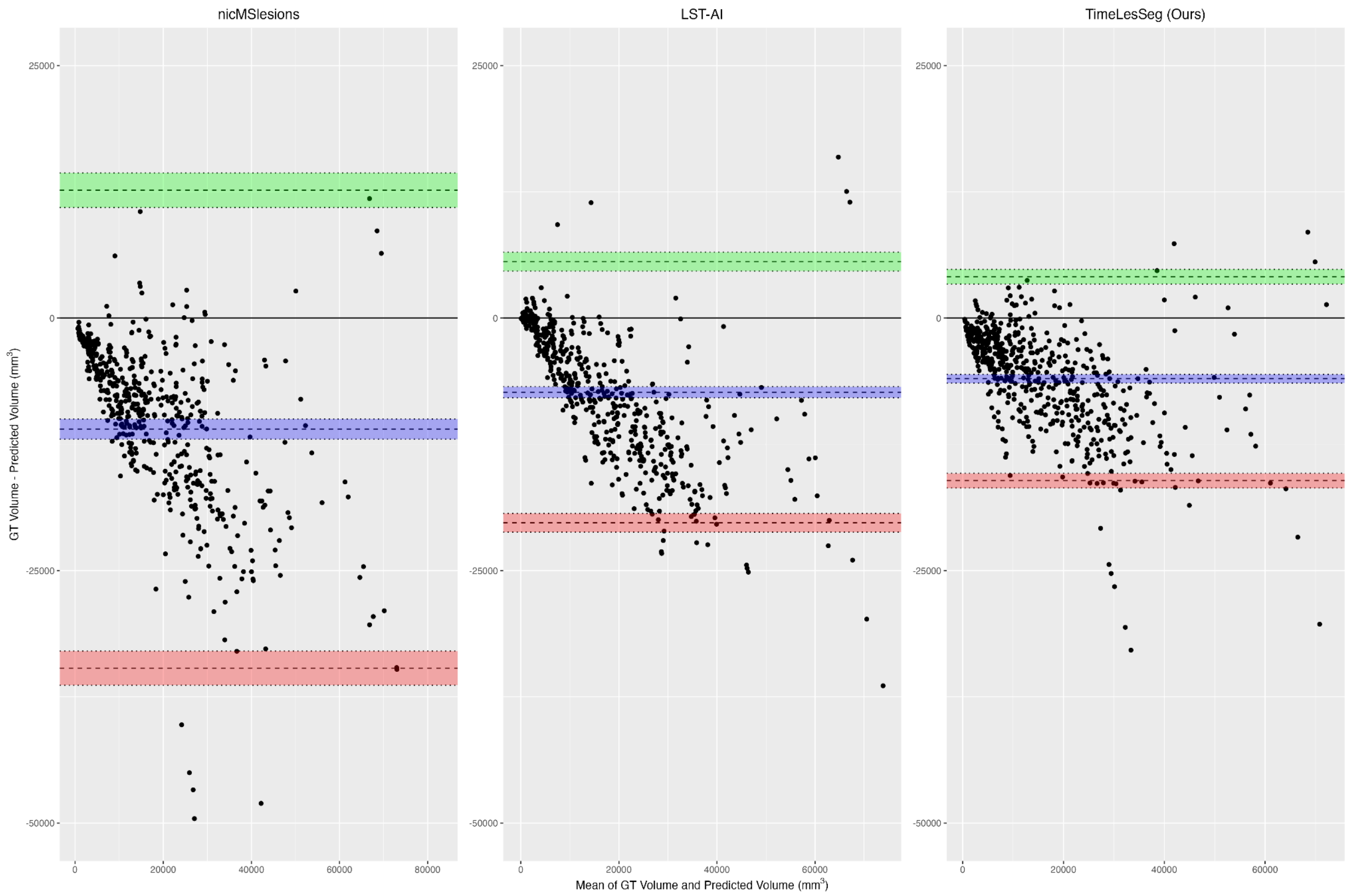

**Figure 5.** *Bland–Altman plots comparing lesion volume agreement between ground truth (GT) and predicted segmentations from MRI scans from 258 people with secondary progressive MS from a clinical trial conducted in the UK (MS-SMART). The plot depicts results for three methods: nicMSlesions (left), LST-AI (middle), and TimeLesSeg (right). The x-axis represents the mean lesion volume (GT and prediction), while the y-axis shows the difference between GT and predicted volumes. The solid black line indicates the mean difference (bias), and the dashed colored lines denote the limits of agreement (±1.96 standard deviations). Each point corresponds to a subject.*

# 4.5 Qualitative results

In this section we showcase qualitative results from all benchmark methods, as well as ours, belonging to subjects from MSLESSEG, Hospital Clínic, and MS-SMART datasets.

On Figure 6, results obtained on FLAIR scans of a subject from MSLESSEG are shown, depicting the comparison between all contrast-agnostic methods: WMHSynthSeg, SAMSEG and TimeLesSeg. As can be observed, WMHSynthSeg tends to “leave holes” within clear, bright hyperintensities, while SAMSEG tends to oversegment periventricular hyperintensities, a bad habit both TimeLesSeg and LST-AI are also prone to (Figure 7). On the other hand, note TimeLesSeg’s sensitivity to small lesions, as it is the only contrast-agnostic method that detects the cluster located in the right midline of subject P19 (Figure 6, top row) – although this leads to a substantial number of false positives.

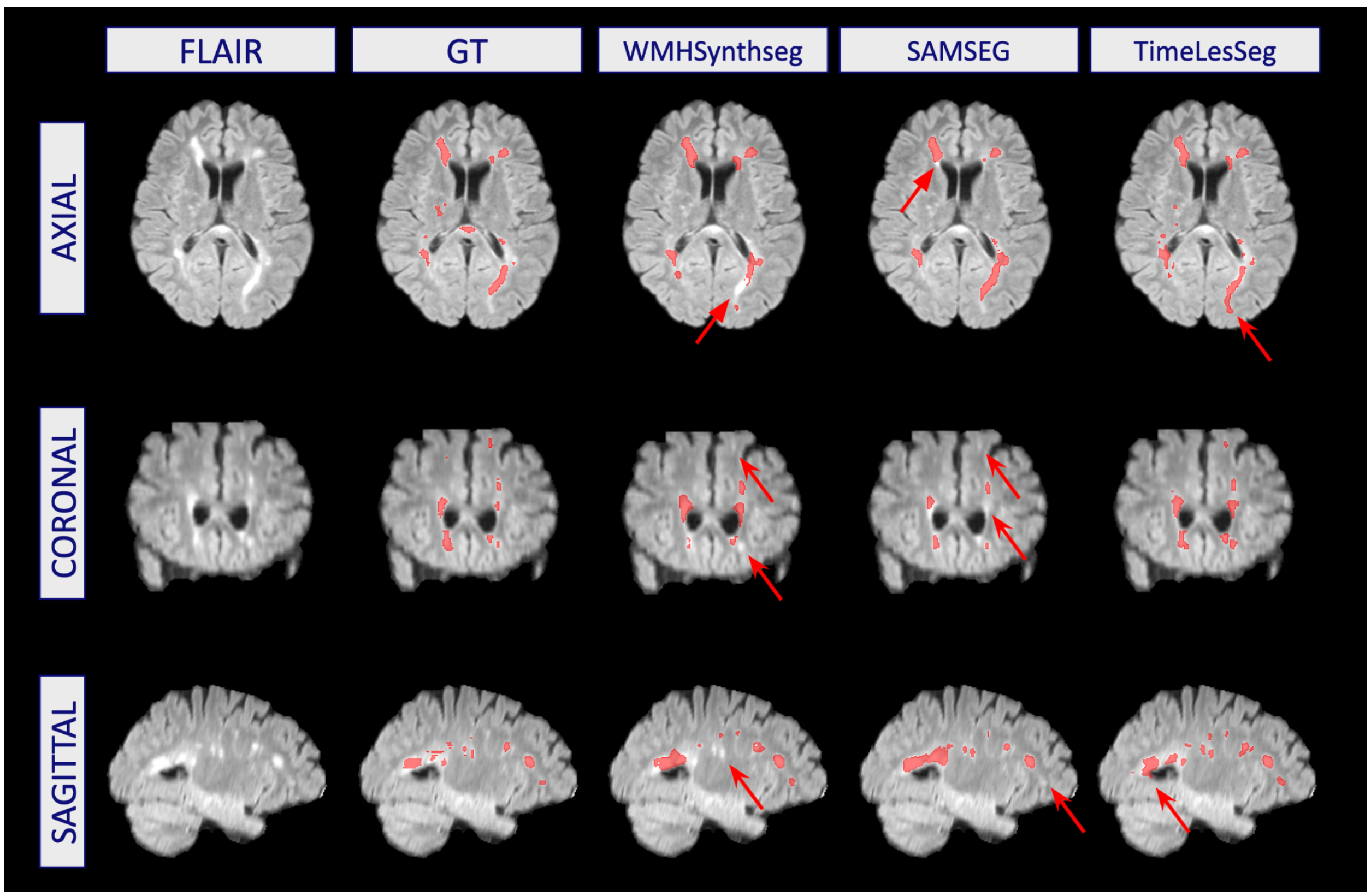

**Figure 6**. *Visualization of slices from all three orthogonal planes of single-modality results from a subject from dataset MSLESSEG (P19's first timepoint, precisely). Each arrow highlights missegmentations – either a missed lesion or a false positive. GT: Ground truth.*

When visually comparing all contrast-agnostic methods against LST-AI, we see that, on the subject depicted in Figure 7, the latter is the only method that correctly segments the small-volume juxtacortical lesion. Note also that TimeLesSeg is the method that over-segments the least normal-appearing periventricular hyperintensities, although this leads to a false negative in Figure 7 (bottom row, right lateral ventricle).

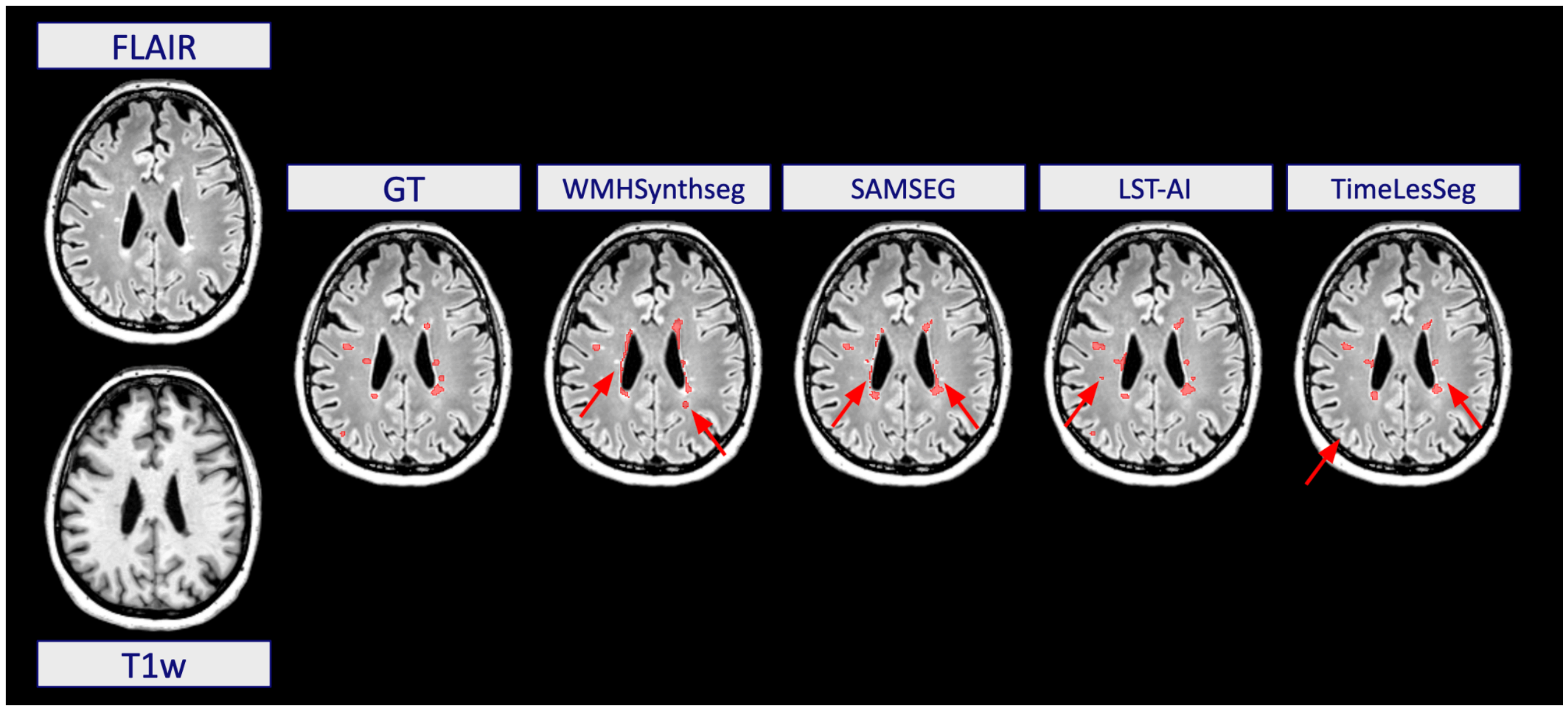

**Figure 7.** *Visualization of axial slices of multi-modality results from a subject from Hospital Clínic's dataset. Each segmentation is overlaid on top of FLAIR volumes. Each arrow highlights missegmentations – either a missed lesion or a false positive. GT: Ground truth.*

Figure 8 displays the results from MS-SMART's clinical trial, where it can be observed that both LST-AI and TimeLesSeg incorrectly detected the CSF pulsation artifact that appears on both lateral ventricles of top row's subject. This artifact, caused by CSF flow dynamics and more commonly found on axially-acquired fast FLAIR images, has been suggested to correlate with aging and ventriculomegaly (Bakshi et al. 2000).

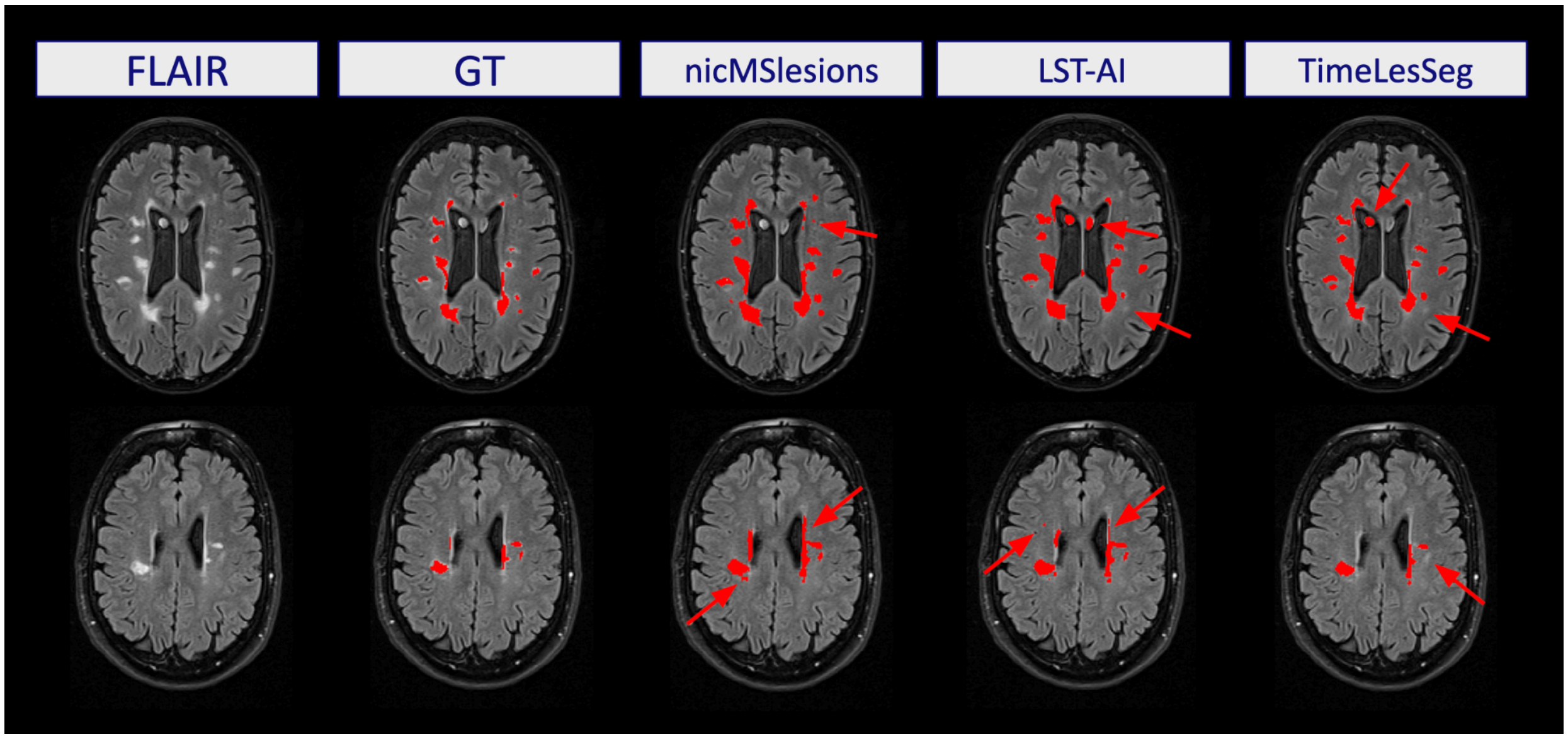

**Figure 8.** *Visualization of axial FLAIR slices of results from two subjects from dataset MS-SMART. Results derived from T1w and FLAIR scans for LST-AI, and single-modality FLAIR scans for both nicMSlesions and TimeLesSeg. Each arrow highlights missegmentations – either a missed lesion or a false positive. GT: Ground truth.*

With TimeLesSeg, longitudinal processing of a sequence of N scans is enabled by providing at each timepoint a lesion mask from a previous one as the second channel of the convolutional neural network model. In the first timepoint, the method functions cross-sectionally – i.e., with an empty lesion mask.

Figure 9 shows an axial slice for all four timepoints of subject one of ISBI dataset (quantitative results in Figure 3, second column). In this case, it can be observed that SAMSEG only segments very bright lesions (on FLAIR scans), and that lesion load decreases with time. On the other hand, we observe that TimeLesSeg tends to oversegment periventricular lesions, and that anatomical changes with time, such as ventricular atrophy, coupled with the strong bias the method has learned where lesions rarely shrink in size (Section 3.1.1), cause an intrusion of lesions towards ventricular CSF, particularly in the last timepoint. Despite that, our method clearly is more precise in detecting lesions (see Figure 4).

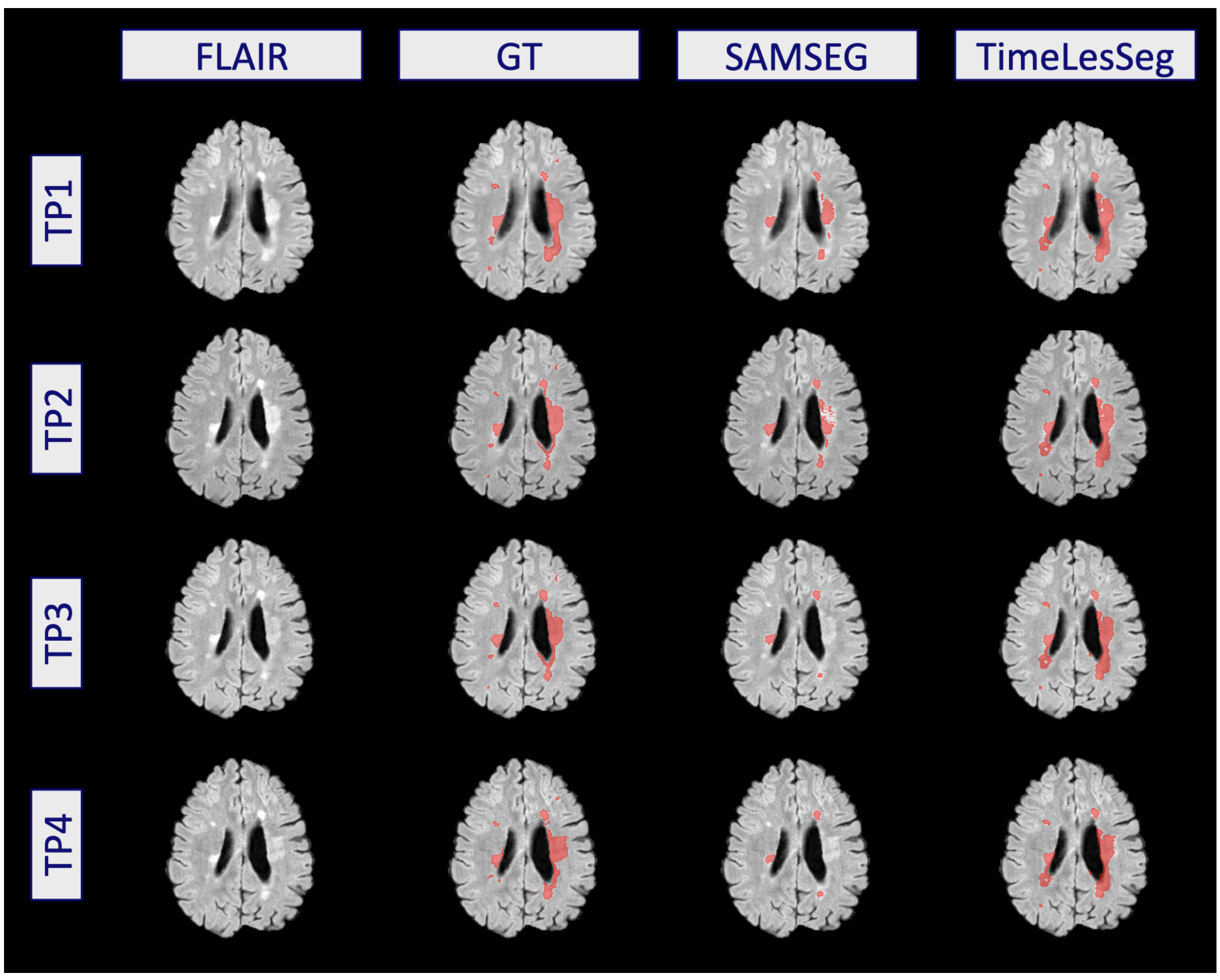

**Figure 9.** *Visualization of axial FLAIR slices belonging to subject 1 of dataset ISBI, across its four timepoints, predicted longitudinally with SAMSEG and TimeLesSeg – the only methods with a longitudinal processing option. GT: Ground Truth; TP1-TP4: Consecutive timepoints.*

# 5-Discussion and concluding remarks

In this work, we have introduced a unified multiple sclerosis lesion segmentation framework that uniformly handles cross-sectional and longitudinal scans using a single model, addressing a long-standing limitation in the field. Existing approaches treat these two settings separately, requiring distinct models, specific acquisition protocols, or explicit temporal image pairs. In contrast, our method generalizes across both temporal and multi-site domains, enabling flexible inference from either a single scan or N input images acquired at arbitrary time points. This flexibility allows the model to operate robustly across heterogeneous longitudinal datasets, even in the presence of changes in scanner hardware, imaging modality, or acquisition artifacts that commonly affect clinical studies.

The ability to use the same model for both segmentation settings has important implications for clinical research and clinical trials, where longitudinal studies frequently suffer from missing time points due to patient dropout, changes in imaging protocols, or incomplete datasets (Gloor et al. 2023). Our approach's seamless adaptation to either setting eliminates the need for ad-hoc solutions, retraining, or separate pipelines, thereby simplifying study design and improving reproducibility and robustness. This versatility also facilitates large-scale retrospective analyses, where temporal completeness cannot be guaranteed.

Our contribution is enabled by a simple conceptual shift. Instead of encoding temporal changes through MRI volumes – inherently susceptible to covariate shifts and noise – we rely solely on the output of their segmentation: lesion masks. As a result, our method is robust to situations where the radiological appearance of lesions evolves, while their boundaries remain consistent – e.g., a stable lesion whose signal intensity changes over time. Using lesion masks as the mechanism to encode priors allows greater flexibility, simultaneously avoiding the pitfalls of intensity-dependent longitudinal comparisons.

Furthermore, modeling the temporal dimension in the binary space helps simplify the task of generating longitudinal data – crucial in conditions like MS where gathering sizable manually-annotated datasets is costly. In our implementation, we have modeled the main patterns of lesion evolution via morphological deformations, enabling the simulation of the time-dependent radiological variability observed in real people with MS. Coupling this approach with a domain-randomization based scan generation method allowed us to synthesize a large number of longitudinal training examples from less than nine parcellation and lesion mask pairs.

This represents a significant departure from contemporary state-of-the-art deep learning techniques such as LST-AI (Wiltgen et al. 2024) and nicMSlesions (Valverde et al. 2017), both of which rely on large, manually annotated datasets for training and development. Our approach instead simplifies the process by modeling lesion evolution in the binary space. This change in conceptual approach has enabled a single convolutional neural network (CNN) to segment any contrast and in both longitudinal and cross-sectional settings, remaining robust to domain shifts and missing modalities, all without the need of ever being exposed to a real scan during training.

In contrast, the other state-of-the-art methods required significantly more manually annotated real data to train. LST-AI was trained on an in-house dataset with 491 T1w and FLAIR pairs. WMHSynthSeg used segmentations from over a thousand subjects and multiple sources (ADNI, ISBI and HCP) to generate its training dataset (Laso et al. 2023). SAMSEG used, on the one hand, a dataset comprised of 212 lesion masks to train the variational autoencoder that models lesion shape; and, on the other, a separate set of 54 pwMS to extend their probabilistic deformable atlas to also model the spatial location of MS lesions (Cerri et al. 2021). Finally, the single-modality version of nicMSlesions that we used in our evaluation experiments was trained on 90 manually segmented 3D FLAIR images. According to their authors, it requires fine-tuning its fully-connected layers on at least a single target image to adapt to new intra-modality domains (Valverde et al. 2019).

Despite requiring orders of magnitude less real data than both competing contrast-agnostic methods and nicMSlesions, our approach performs significantly better on single-modality inputs (Figure 3), better than WMHSynthSeg and on par with SAMSEG on multimodal inputs (Figure 4), and better than nicMSlesions in Dice Similarity Coefficient (DSC) and Average Symmetric Surface Distance on the dataset from the MS-SMART clinical trial (Figure 6).

One potential explanation for the proposed method’s drop in Dice Similarity Coefficient on multimodal inputs is that it predicts each modality independently, ensembling the result by averaging their probabilities to provide the final segmentation. This leads to cross-modality information being aggregated at a much later stage compared to SAMSEG or LST-AI, as no crosstalk between features extracted from each modality occurs until after their voxel-wise probabilities have been derived. In comparison, SAMSEG models all scan modalities simultaneously with a single GMM-based probability density function, allowing for a joint optimization process.

On the other hand, as Figure 4 clearly demonstrates, LST-AI currently outperforms all contrast-agnostic methods when applied to research data of the same modalities that it was trained on (FLAIR and T1w). This contrast-specific method processes its two-channel input with three separate CNNs and subsequently averages their output lesion probabilities. Each CNN extracts features on both modalities simultaneously, aggregating them from the first convolutional layer. This might aid in providing more nuanced and comprehensive semantic information by explicitly combining the distinct biological properties of pathological tissue each modality highlights. However, its application relies on the assumption that all subjects will consistently have T1w and FLAIR modalities available, which might not hold in longitudinal clinical trials, research studies, nor clinical settings. Furthermore, its rigidity precludes its application on longitudinal inputs, treating temporal data points from the same subject as independent from each other.

Therefore, a potential area of improvement would be explicitly combining cross-modality information during feature extraction. Multi-modal deep learning approaches robust to the issue of missing modalities are best exemplified by HeMIS’ framework (Havaei et al. 2016), where all available modalities are encoded independently by two-convolutional-layer deep encoders, and the resulting features are summarized into first and second order statistics – which are independent from the number of modalities –, that are then decoded into a segmentation. Noting the inherent limitation of mean and variance for encoding the cross-modality anatomical dependencies, a more recent approach utilized a multi-modal Variational Auto-Encoder (MVAE) to model all modality-specific information into a shared latent space through conditional independence (Dorent et al. 2019). Although the translation to a contrast-agnostic paradigm is not straightforward, these serve as potential starting points for future research.

Secondly, it remains unexplored whether training our unified approach on a combination of synthetic and real scans, as previously described (Goebl et al. 2025), would yield performance on par with contrast-specific methods. Nonetheless, we pose that this would contradict our current domain randomization paradigm where a neural network is exposed to maximal variability during training by relying on synthetic, unrealistic data.

It must be also noted that, stemming from our mask-reliant paradigm, and all too common in GPT-based Natural Language Processing (NLP), our CNN behaves in an *autoregressive* manner, i.e., predicts future outputs (segmentations) conditioned on past ones (see Figure 1). In both cases, issues can arise from the inherent discrepancy between training and inference processes. During training, the model learns to produce outputs from high-quality inputs – human-written text in NLP or, in our case, controlledly augmented lesion masks where only a subset of lesions has been deformed, while the rest perfectly align with the scan (stable lesions). However, during inference, the method can only base future segmentations on its own predictions, leading to error compounding and a deviation from biologically plausible lesion trajectories.

This phenomenon – autoregressive error aggregation – has been widely studied in time-series forecasting (Parthipan et al. 2024), yet it remains an open challenge. While extreme instability is easily detectable, more subtle drifts – where predictions remain anatomically coherent yet gradually diverge from true lesion evolution – are harder to identify. Nevertheless, understanding and formally quantifying autoregressive error accumulation in longitudinal neuroimage processing remains an important avenue for future work. Potential approaches to explicitly tackle this issue include randomly injecting noise to baseline masks during training (Pasini et al. 2024); or scheduled sampling, a training technique where the network is increasingly exposed to its own predictions (Bengio et al. 2015).

In line with the last point, increasing noise thus deviating from modeling realistic lesion evolution patterns in the last application of *FLM* in our generative pipeline (see Section 2.4 and Figure 2) might aid in enhancing our method's robustness to its own errors during inference, alleviating error compounding. In any case, it must be noted that, in the experiments described in this manuscript, *FLM*'s parameters were chosen heuristically to assess initial feasibility of the approach – no optimization was performed –, leaving room for a more detailed analysis. We also leave for future work the optimization of the weighting factors for each modality, which was not systematically explored in this manuscript – equal per-modality weights were used in the experiments here presented. A more rigorous fine-tuning strategy might further improve segmentation performance.

In conclusion, we have presented a unified framework for multiple sclerosis lesion segmentation that consistently addresses both cross-sectional and longitudinal data within a single, integrated model. Our work also introduces a novel paradigm for dataset generation in lesion segmentation research. Instead of synthesizing images or explicitly modelling the biological evolution of disease, our retrospective approach generates lesion masks that span a plausible range of spatial and morphological variability, from the starting point of the disease till the current acquisition. This approach decouples lesion shape evolution from image appearance, enabling the training of robust segmentation networks without relying on large quantities of realistic synthetic images or complex temporal disease models. We believe this direction opens new opportunities for data-efficient and domain-agnostic training strategies, especially in settings where real annotated datasets are limited.